\pdfoutput=1

\documentclass[11pt]{article}

\usepackage{acl}

\usepackage{times}
\usepackage{latexsym}
\usepackage{xcolor}
\usepackage{tabularx}
\usepackage{booktabs}

\usepackage[T1]{fontenc}

\usepackage[utf8]{inputenc}

\usepackage{microtype}

\title{Xu at SemEval-2022 Task 4: Pre-BERT Neural Network Methods vs Post-BERT RoBERTa Approach for Patronizing and Condescending Language Detection}

\author{ Jinghua Xu\\
 University of Tübingen \\
 \texttt{jinghua.xu@student.uni-tuebingen.de}\\}
 
\begin{document}

\maketitle

\begin{abstract}

This paper describes my participation in the SemEval-2022 Task 4: Patronizing and Condescending Language Detection. I participate in both subtasks: Patronizing and Condescending Language (PCL) Identification and Patronizing and Condescending Language Categorization, with the main focus put on subtask 1. The experiments compare pre-BERT neural network (NN) based systems against post-BERT pretrained language model RoBERTa. This research finds that NN-based systems in the experiments perform worse on the task compared to the pretrained language models. The top-performing RoBERTa system is ranked 26 out of 78 teams (F1-score: 54.64) in subtask 1, and 23 out of 49 teams (F1-score: 30.03) in subtask 2.

\end{abstract}

\section{Introduction}

An entity is considered to engage Patronizing and Condescending Language (PCL) when its language use presents a superior attitude towards others or depicts them in a compassionate way \cite{perezalmendros2020dont}. Such language is often used toward vulnerable communities such as women, refugees, and homeless people. These unfair treatments of the vulnerable groups are believed to result in further exclusion and inequalities in society. Compared to other types of harmful language (e.g. hate speech), PCL is considered more subtle and unconscious. Given the negative effects of PCL on society and its subtle nature, enabling computers to identify and categorize PCL presents an interesting technical challenge to the NLP community.

This paper describes my participation in both subtasks of the SemEval-2022 Task 4: Patronizing and Condescending Language Detection \citep{perezalmendros2022semeval}. Subtask 1, Patronizing and Condescending Language Identification is a binary text classification task to predict whether a given paragraph contains PCL or not. Subtask 2, Patronizing and Condescending Language Categorization is a multi-label classification task to identify the categories of a given paragraph according to the taxonomy defined in \citet{perezalmendros2020dont}, which categorizes PCL into 7 types: \textit{1) Unbalanced power relations 2) Shallow solution 3) Presupposition 4) Authority voice 5) Metaphor 6) Compassion 7) The poorer, the merrier}. The dataset \citep{perezalmendros2020dont} contains annotated paragraphs in English, collected from news stories in 20 English-speaking countries.\footnote{19 countries and Hong Kong: Australia, Bangladesh, Canada, Ghana, Ireland, India, Jamaica, Kenya,
Sri, Lanka, Malaysia, Nigeria, NewZealand, Philipines, Pakistan, Singapore, Tanzania,
UK, United States, South Africa, and the special administrative region of China, Hong Kong.}

The focus of my experiments is primarily on subtask 1, meanwhile, this paper also proposes a solution to subtask 2. For subtask 1, the experiments compare pre-BERT neural network (NN) based systems including a majority voting system of NN models against pretrained language model RoBERTa. The experiments start with building individual NN models from the most basic artificial neural network (ANN) to long short-term memory network (LSTM) models following previous work on NN for text classification. It was found that the NN-based systems in the experiments perform worse on this task in comparison to the pretrained language models. The best-performing NN-based voting system could not outperform the RoBERTa baseline model. For subtask 2, this paper simply proposes a RoBERTa solution.

The code is released at: \href{https://github.com/JINHXu/PCL-Detection-SemEval2022-task4}{github.com/JINHXu/\\PCL-Detection-SemEval2022-task4}.


\section{Background}

Numerous previous research had been conducted on the treatment of condescension and patronization. The studies range in various areas from sociolinguistics \citep{irisawa1993cardiac} to medicine \citep{komrad1983defence}. Whereas in the field of natural language processing, automatically identifying or categorizing PCL has been an understudied area. Most research on harmful language detection has focused on more explicit and aggressive topics such as hateful speech \citep{macavaney2019hate}, offensive language \citep{zampieri2019semeval}, and fake news \citep{conroy2015automatic}. Only in recent years, few in the research community have started to show interest in enabling computers to identify condescending language. Prior to this shared task, for instance, \citet{wang2019talkdown} proposed an annotated TalkDown corpus of condescending language from social media.

\section{Dataset}

The corpus used for this shared task, the Don't Patronize Me! dataset is described in \citet{perezalmendros2020dont}. The corpus consists of 10,637 paragraphs extracted from the News on Web (NoW) corpus \citep{davies2013corpus}. The paragraphs were selected according to 10 keywords related to potentially vulnerable communities (disabled, homeless, hopeless, immigrant, in need, migrant, poor families, refugee, vulnerable, and women). And for each keyword, a similar number of paragraphs were chosen for each of the 20 English-speaking countries.\footnote{Except for Hong Kong, which is not a country.} Each paragraph is annotated with a true/false label indicating whether it contains PCL or not, and the ones that contain PCL are annotated with a category label. Each category label is a set of 7 binary predictions, each prediction indicates the existence of a specific type of PCL.

The dataset is highly imbalanced. The POS:NEG ratio is approximately 1:10, which poses a challenge to the predictive modeling process. In the experiments of this paper, various strategies were employed in order to deal with the imbalance in data. The following sections will describe these strategies in detail.

Additionally, the shared task provides a comparable 80/20 split of the training data for development. In the experiments of this paper, the same split was used to train models and generate predictions in the development stage.


\section{Model}




\subsection{Preprocessing}

In the preliminary experiments, it was found that preprocessing data by removing stop words decreases model performance. Thus in the following model training process, the text data are used as-is. Furthermore, in order to deal with data imbalance, various approaches including data oversampling, undersampling, and setting class weights were experimented with. The neural network models were found in preliminary experiments to work the best with the original data, with class weights set to 10:1 according to the POS:NEG ratio in data. Whereas the RoBERTa models present the best performance with oversampled data with default class weights.

\subsection{Neural Network Models}

The experiments start with exploring NN models for the binary text classification subtask. Previous work has shown that linear classic machine learning models such as Linear SVM \citep{suthaharan2016support}, Bernoulli Naive Bayes \citep{webb2010naive}, and Logistic Regression \citep{wright1995logistic} have advanced performance on binary text classification with proper feature engineering. This paper is, however,  interested in exploring neural network solutions to the task, given the sufficient size of the dataset. 

Neural network models have been regarded to be capable of achieving remarkable performance on text classification. In addition to the popular LSTM \citep{hochreiter1997long} models, some basic ANN \citep{mcculloch1943logical} models have also been proved in previous work to perform well on the task of binary text classification. The experiments of this paper start with building individual NN models from the most basic ANN models to the more sophisticated LSTM models. Furthermore, in order to continue improving system performance from the individual models, a majority voting system that uses the predictions of both of the best-performing ANN and LSTM models was built. 


\subsubsection{Basic ANNs}
 
Common basic ANN architectures for binary text classification tasks typically consist of an Embedding layer, a pooling layer of different types (average, minimum, maximum), and various dense layers. Following the previous work, the experiments start with building a baseline ANN model using a GloVe \citep{pennington2014glove} embedding layer for word representation, with a \texttt{GlobalAveragePooling1D} built on top of it, followed by a ReLU layer, and a sigmoid layer to generate predictions. In the preliminary experiments, various types of pooling were tried, and the model with the \texttt{GlobalAveragePooling1D} layer presented the best performance. The confidence threshold was initially set to 0.5, which resulted in low precision and high recall. Thus the threshold was gradually increased in experiments, with 0.7 found to generate the best predictions in the development stage.

In order to improve the ANN model from the baseline, more dense layers were added to the network gradually. Since there are no rules of thumb in building a neural network, the strategy employed in this experiment is to continue adding dense layers of tanh and ReLU to the baseline model before the output layer. The F-score reaches a peak value after two additional tanh layers with a ReLU layer in between were added to the baseline model. Adding more dense layers did not further help increase the model performance in the experiments. 

\subsubsection{LSTM}

The LSTM model uses the same GloVe embedding layer for word representation, with a single layer of LSTM units (output dimension size  60) built on top of it. A \texttt{GlobalMaxPool1D} layer is built on top of the LSTM layer, followed by a ReLU layer, and a sigmoid layer to generate predictions. The dropout rate is set to 0.1. With the confidence threshold set to 0.5, relatively even precision and recall were obtained.

\subsubsection{NN voting system}

NN model performance can be unstable in each run, this was also confirmed in the experiments of this paper. In order to handle the instability, also to continue increasing system performance from the individual models, a majority voting system based on both NN models was built. The system considers the predictions of both the ANN and the LSTM models in two separate runs, which results in four votes in total. The systems prediction for each paragraph is then based on the majority vote of the four votes produced by both models in two runs.\\

All NN models in the experiments are implemented using \texttt{tensorflow.keras} \citep{chollet2015keras}. During training, each model uses 10\% of data for validation, with class weights set to 10:1 as mentioned in a previous section.

The hyperparameter tuning process in this experiment focuses on batch size and the number of training epochs. For each model, batch sizes of 16, 32, 64, 128, and training epochs of 10, 50, 100 were tried. All models present the best performance with the number of training epochs set to 50. The LSTM model works the best with training batch size of 128, and ANN models with 32.

\subsection{Pretrained Language Model: RoBERTa}

For both shared tasks, this paper proposes a RoBERTa \citep{liu2019roberta} solution. RoBERTa is regarded as an improved pretraining procedure from BERT \citep{devlin2018bert}, and it is able to match or exceed the performance of all post-BERT methods. All pre-trained language models in the experiments are implemented using the \texttt{simpletransformers} library \citep{rajapakse2019simpletransformers}.

In subtask 1, the shared-task provides a baseline \texttt{roberta-base} model with default configurations, trained on undersampled data. On top of this work, I further tuned the hyperparameters (mainly the number of training epochs, in the search space: 1, 2, 3, 5, 10) and improved model configurations using manually oversampled/undersampled PCL data of various POS:NEG ratios, class weights for fine-tuning. In addition to the \texttt{roberta-base} model used in the baseline model, I also experimented with a number community models\footnote{A list of community models can be found on the website: \href{https://huggingface.co/models}{huggingface.co/models}.} alternative to \texttt{roberta-base}. Among the community models, the experiments mainly focus on BERT- and RoBERTa-based models for toxic language detection and sentiment analysis, given the similarity and relevance of the tasks to PCL detection. Appendix \ref{sec:appendix} lists the community models tried in the experiments. However, none of these community models in Appendix \ref{sec:appendix} turned out to work better than \texttt{roberta-base} in my experiments. I believe the models are too specialized in their own tasks (e.g. sentiment analysis, toxic language detection), therefore resulting in poor performance on the PCL detection task. 


The best-performing model in the development stage is a \texttt{roberta-base} model, with the number of training epochs set to 1, maximum input sequence length increased to 500.\footnote{The longest paragraph in training data is of the length between 400 and 500, in case of future longer data instances to predict on, the model's maximum input sequence length is set to 500.} The data was balanced by manually repeating all positive data instances 9 times, and keeping the same number of negative data instances for training. The balanced dataset results in 8937 data instances of each class for training. In order to reduce training time, GPUs were used for model training and inference. All RoBERTa-related experiments were conducted on Google Colab.\footnote{\href{https://colab.research.google.com/}{colab.research.google.com/}}


In subtask 2, the shared-task also provides a RoBERTa baseline model configured with \texttt{roberta-base} model, with default configurations. The baseline model is trained on undersampled data (794 positive data instances, 397 negative data instances). I simply increased the maximum input sequence length to 500 from the baseline model and oversampled data to obtain 7146 positive data instances and 7146 negative data instances for training.

\section{Results}

\subsection{Subtask 1: PCL Detection}

\begin{table}[h]
\begin{tabularx}{\columnwidth}{lccc}
\hline
 Model & Precision & Recall & F-score \\
\hline
ANN\textsuperscript{baseline}&32.63&39.19&35.61\\
ANN & 36.50&46.23&40.79\\
LSTM&41.44&46.23&\textbf{43.70}\\
voting\_NN&46.29&40.70&43.32\\
\hline
RoBERTa\textsuperscript{baseline}&40.98&50.25&45.14\\
RoBERTa-base&51.15&66.83&\textbf{57.95}\\
BERT-emotion&35.93&41.7&38.6\\
RoBERTa-toxic&37.5&66.33&47.91\\
\hline
\end{tabularx}
\caption{Model performance on development data.}\label{Table 1}
\end{table}

Table \ref{Table 1} shows the precision, recall, and F1-score of the models in the development stage. Among the neural network systems, the LSTM model presents the most advanced performance with an F-score of 43.7. Meanwhile, the voting system has an F-score (43.32) only slightly lower than the LSTM model. The F-score of the ANN model is lower than the LSTM model, however, the performance gap is not huge. Nevertheless, the best-performing neural network based system LSTM does not outperform the RoBERTa baseline model.

The tuned RoBERTa-base model has the highest score of 57.8 among all systems in the development stage. As mentioned in a previous section, the community models pretrained on sentiment or toxicity language data present poor performance on the PCL data compared to the base model of RoBERTa.\footnote{Only the performance of two of the community models tried in the experiments are presented in the tables.} Overall, for each model in the development stage, the difference between precision and recall is not vast, except for the tuned RoBERTa model and the toxicity model of RoBERTa.

\begin{table}[h]
\begin{tabularx}{\columnwidth}{lccc}
\hline
 Model & Precision & Recall & F-score \\
\hline
ANN\textsuperscript{baseline}&28.34&48.90&35.88\\
ANN&26.62&67.19&38.14\\
LSTM&38.31&50.16&43.44\\
voting\_NN&48.50&40.69&\textbf{44.25}\\
\hline
RoBERTa\textsuperscript{baseline}&39.02&62.78&48.13\\
RoBERTa-base&46.19&66.88&\textbf{54.64}\\
BERT-emotion&36.54&35.96&36.25\\
RoBERTa-toxic&25.19&84.54&38.81\\
\hline
\end{tabularx}
\caption{Model performance on test data.}\label{Table 2}
\end{table}


Table \ref{Table 2} presents model performance on the test data in the evaluation stage. It is notable that among the neural network models, the top-performing system on test data becomes the voting system (F-score: 44.25) instead of the LSTM model, which performs the best in the development stage. The F-score of the voting system in the evaluation stage is also higher than in the development stage. Nonetheless, as the top-performing neural network based system in the evaluation stage, the NN voting system presents an F-score still lower than that of the RoBERTa baseline model (F-score 48.13). Both the LSTM and the ANN model F-score decreased from in the development stage. While the baseline ANN model presents a similar F-score on the test data to that on the development data. In general, in the evaluation stage, the gap between precision and recall is rather big for both the ANN baseline and the ANN model, whereas it is small for the LSTM model and the NN voting system. By comparing the performance of the neural network based systems during the development stage to the evaluation stage, it can be seen that the performance of the ANN models is less stable compared to the LSTM and the voting system.

The tuned RoBERTa-base model is still the top performer among all models in the evaluation stage, with an F-score of 54.64. However, this score decreased from in the development stage. While for the RoBERTa baseline model, the F-score in the evaluation stage is higher than in the development stage. As for the two community models, their performance on test data is also worse than on the development data. Overall, every RoBERTa model shows higher recall than precision with a notable gap. This is also true for the ANN models as mentioned in the previous paragraph. These models produce more false positives than false negatives. While for the BERT-based emotional model, it shows similar precision and recall, although also a low F-score. In general, for every model in the evaluation stage except for the BERT model, the gap between precision and recall further increased from that in the development stage. 

\subsection{Subtask 2: PCL Categorization}

\begin{table}[h]
\centering
\begin{tabularx}{\columnwidth}{lcc}
\toprule
F-score&RoBERTa\textsuperscript{baseline}& RoBERTa\\
\midrule
Unb. power rel. & 35.35 & 55.94\\
Shallow solu. & 00.00 &31.74\\
Presupposition & 29.63 &24.44\\
Authority voice. & 00.00 & 	19.35\\
Metaphor & 00.00 & 23.88\\
Compassion & 28.78 & 	45.83\\
The p., the mer. & 00.00 & 15.38\\
Average & 13.40 & 30.94\\
\bottomrule
\end{tabularx}
\caption{Model performance on development data.}\label{Table 3}
\end{table}

Table \ref{Table 3} presents the per-class and average F-scores of the RoBERTa baseline model and the proposed RoBERTa model for subtask 2 in the development stage. The proposed RoBERTa model is able to produce a higher F-score for each class with the exception of the \textit{Presupposition} category. Overall, the average F-score is improved from baseline by around 17\%.

\begin{table}[h]
\centering
\begin{tabularx}{\columnwidth}{lcc}
\toprule
F-score&RoBERTa\textsuperscript{baseline}& RoBERTa \\
\midrule
Unb. power rel. & 35.35 & 54.38 \\
Shallow solu. & 00.00 & 47.06 \\
Presupposition & 16.67 & 26.92 	\\
Authority voice. & 00.00 &	24.06 	\\
Metaphor & 00.00 &	11.11 	\\
Compassion & 20.87 & 46.72 \\	
The p., the mer. & 00.00 & 00.00 \\
Average & 10.41 & 30.03\\
\bottomrule
\end{tabularx}
\caption{Model performance on test data.}\label{Table 4}
\end{table}

Table \ref{Table 4} presents the per-class and average F-scores of the RoBERTa baseline model and the proposed RoBERTa model for subtask 2 in the evaluation stage. As can be seen from the table, the proposed RoBERTa model increased the per-class F-score from the baseline model for each category except for only the \textit{the poorer the merrier} class, for which neither the baseline model nor the proposed model is able to detect. The average F-score of the proposed model is also increased from that of the baseline model. However, the score slightly decreased in the evaluation stage from in the development stage.

\section{Conclusion}

The experiments of this paper compare some of the pre-BERT neural network based systems against the post-BERT pretrained language model RoBERTa. The experiments start with building individual NN models from the most basic ANN models to the more sophisticated LSTM models, and create a majority voting system based on the individual NN models. It was found that the NN-based systems in the experiments perform worse on the task compared to the RoBERTa baseline model. And the community models pretrained on relevant data such as sentiment and toxicity data turn out to be too specialized in their own task thus resulting in poor performance on the PCL data compared to the RoBERTa-base model.

This paper explores neural network models mainly the basic ANN and LSTM models. Future work should also consider convolutional neural networks (CNN) for PCL detection. In addition to the neural networks, I suggest also investigating classic machine learning models such as Logistic Regression, as well as indicative linguistic features of PCL for feature engineering. Furthermore, on top of the RoBERTa-base model, I propose to pretrain a RoBERTa model using the TalkDown corpus proposed in \citet{wang2019talkdown}, and fine-tune the pretrained model using the PCL data. Finally, future work should run further error analysis of the models to improve performance.

\section*{Acknowledgements}

I thank Dr. Çağrı Çöltekin for his guidance and help throughout the process. 

\bibliography{anthology,custom}

\appendix

\section{Appendix: List of Community Models}
\label{sec:appendix}
mohsenfayyaz/toxicity-classifier\\
mohsenfayyaz/roberta-base-toxicity\\ 
SkolkovoInstitute/roberta\_toxicity\_classifier\\
mohsenfayyaz/toxicity-classifier\\
DaNLP/da-bert-emotion-binary \\
siebert/sentiment-roberta-large-english\\ 

\end{document}